# SIMAX: A Scalable and Interpretable Framework for Multi-Fidelity and Annotated Clinician–Patient Dialogue Simulation

Zhuhan Bao[1†], Rui Yang[2,3†], Bohao Yang[4], Zhiyi Liu[1], Sicheng Shu[1], Ruio Heerschap[1,5], Le Li[6], Doris Yang[7], Elisabeth Bond[1], Haoyuan Wang[8,9], Nicoleta Economou-Zavlanos[1], Joshua M. Biro[10], Matthew McDermott[11], Nan Liu[1,2,3,15,16], Anand Chowdhury[17], Kai Sun[14], Kathryn Pollak[12,13], Ed Hammond[18], Chuan Hong[1,19*]

[1] *Department of Biostatistics and Bioinformatics, Duke University School of Medicine, Durham, NC, USA*

[2] *Duke-NUS AI + Medical Sciences Initiative, Duke-NUS Medical School, Singapore, Singapore*

[3] *Centre for Biomedical Data Science, Duke-NUS Medical School, Singapore, Singapore*

[4] *Department of Statistical Science, Duke University, Durham, NC, USA*

[5] *Leiden University Medical Centre, Leiden, The Netherlands*

[6] *Department of Mathematics, University of Texas at Austin, Austin, USA*

[7] *Department of Internal Medicine, Yale School of Medicine, New Haven, CT, USA*

[8] *Department of Biostatistics, Epidemiology and Informatics, Perelman School of Medicine, University of Pennsylvania, Philadelphia, PA, USA*

[9] *The Graduate Group in Applied Mathematics and Computational Science, School of Arts and Sciences, University of Pennsylvania, Philadelphia, PA, USA*

[10] *Medstar Health National Center for Human Factors in Healthcare, Washington, DC, USA*

[11] *Department of Biomedical Informatics, Columbia University, New York, NY, USA*

[12] *Cancer Prevention and Control, Duke Cancer Institute, Durham, NC, USA*

[13] *Department of Population Health Sciences, Duke University School of Medicine, Durham, NC, USA*

[14] *Division of Rheumatology and Immunology, Duke University School of Medicine, Durham, NC, USA*

[15] *Pre-hospital and Emergency Research Centre, Health Services Research and Population Health, Duke-NUS Medical School, Singapore, Singapore*

[16] *NUS Artificial Intelligence Institute, National University of Singapore, Singapore, Singapore*

[17] *Division of Pulmonary, Allergy and Critical Care Medicine, Duke University School of Medicine, Durham, NC, USA*

[18] *Duke Center for Health Informatics, Duke University, Durham, NC, USA*

[19] *Duke Clinical Research Institute, Durham, NC, USA*

[†]*: Zhuhan Bao and Rui Yang contributed equally*

**Correspondence to: Chuan Hong, Email: chuan.hong@duke.edu*

## ABSTRACT

**Background.** The widespread deployment of ambient digital scribes is driving the large-scale capture of clinician–patient dialogues. Traditional approaches that rely on human coders to annotate clinical communication data struggle to meet the demands of efficiency, consistency, and scalability, making AI-driven communication coding systems increasingly important. However, evaluating such systems depends on real-world clinical dialogues and human-coded labels, both of which are difficult to obtain at scale.

**Methods.** We developed SIMAX (**S**calable and **I**nterpretable Framework for **M**ulti-Fidelity and **A**nnotated Clinician–Patient Dialogue **S**imulation), a framework for generating large-scale, controlled clinical dialogue data with reference behavioral annotations. SIMAX generates clinician–patient dialogues based on predefined clinical scenarios, personas and voice conditions, and target communication behaviors. Communication behaviors are controlled using two complementary codebooks: the Global Codebook for overall communication quality and the WISER Codebook for specific and countable communication behaviors. We evaluated SIMAX across intrinsic data quality and downstream utility, including automated and human quality assessments and an example evaluation of a communication coding system.

**Results.** SIMAX generated 3,388 simulated dialogues covering three clinical specialties, multiple visit stages, persona characteristics, and accent conditions. Automated assessment showed mean UTMOS and WV-MOS scores of 3.03 and 2.61, suggesting reasonable speech naturalness; WER and CER were 0.07 and 0.05, indicating high transcription fidelity; and CLAP cosine similarity was 0.41, indicating positive text–audio semantic correspondence. Human evaluation showed a median MOS of 4.67, suggesting good clarity, naturalness, and speaker differentiation, and a median clinical realism score of 3.00. Downstream evaluation suggests that SIMAX generated data can preliminarily assess how the communication coding system responds to different behavioral targets and identifies insufficient sensitivity in certain dimensions.

**Conclusions.** SIMAX generates controlled and reproducible simulated clinician-patient dialogues, providing a data foundation for the development, validation, and refinement of communication coding systems.

## Introduction

Ambient artificial intelligence (AI) systems, particularly automated digital scribes (ADS), are increasingly being introduced into routine clinical practice[1,2]. These systems are primarily designed to reduce documentation burden, improve workflow efficiency, and allow clinicians to focus more directly on patients[3]. At the same time, the widespread deployment of ADS means that large volumes of clinician–patient dialogues can be captured and transcribed, generating clinical communication data at an unprecedented scale[4,5]. As these data continue to grow, traditional approaches that rely on human coders to annotate communication behaviors record by record will struggle to meet the demands of efficiency, consistency, and scalability[6]. Therefore, AI-driven communication coding systems capable of automatically analyzing clinical communication data are becoming increasingly important[7,8].

However, rigorous evaluation of AI-driven communication coding systems remains challenging. These systems are typically assessed by comparing their outputs against human-coded labels to determine whether they align with established coding standards. One key challenge is that the collection, sharing, and reuse of real clinician–patient dialogues are constrained by privacy and regulatory requirements[9], making existing data insufficient to cover diverse clinical settings, patient populations, and edge-case communication scenarios. Another key challenge is that high-quality human labels are difficult to obtain. Human annotation depends on trained coders, making the process time-consuming and difficult to scale[10]. Together, these barriers create a critical gap: large-scale, privacy-compliant clinician–patient dialogue datasets with reliable behavioral labels remain scarce, limiting the systematic evaluation of communication coding systems.

Synthetic clinician–patient dialogues offer a feasible complementary approach for constructing evaluation data[11,12]. However, existing studies have largely prioritized content realism, focusing on whether generated dialogues are consistent with underlying clinical records, while placing less emphasis on the controllability of interactional behaviors[13]. For evaluating communication coding systems, dialogues that are merely realistic in content are not enough; ideal evaluation data should include explicit behavioral targets as well as reliable reference labels[8]. Therefore, existing synthetic dialogue resources remain limited in their ability to support controlled and reproducible evaluation of communication coding systems.

To fill this gap, we developed SIMAX (**S**calable and **I**nterpretable Framework for **M**ulti-Fidelity and **A**nnotated Clinician–Patient Dialogue **S**imulation), a simulation framework for generating large-scale clinician–patient dialogues. The value of SIMAX lies in providing actively constructible data, enabling communication coding systems to be tested reproducibly under predefined communication behavior targets rather than relying only on limited samples of real-world dialogues. By covering diverse clinical contexts, patient characteristics, and communication behaviors, SIMAX enables users to generate clinical dialogues for the scenarios they need, thereby supporting the broader development, validation, and refinement of communication coding systems.

## Methods

### SIMAX Framework

*Overall Workflow*

As shown in **Figure 1**, SIMAX builds on predefined experimental configurations to generate behaviorally controlled clinician–patient dialogues. The experimental configuration defines the clinical scenario, persona and voice conditions, and target communication behaviors for each simulation. The target communication behaviors are translated through structured communication codebooks into concrete behavioral definitions, which are used to guide text dialogue generation. The generated text dialogue is subsequently converted into two-speaker audio and stored together with behavioral target annotations and related metadata, yielding a multimodal record.

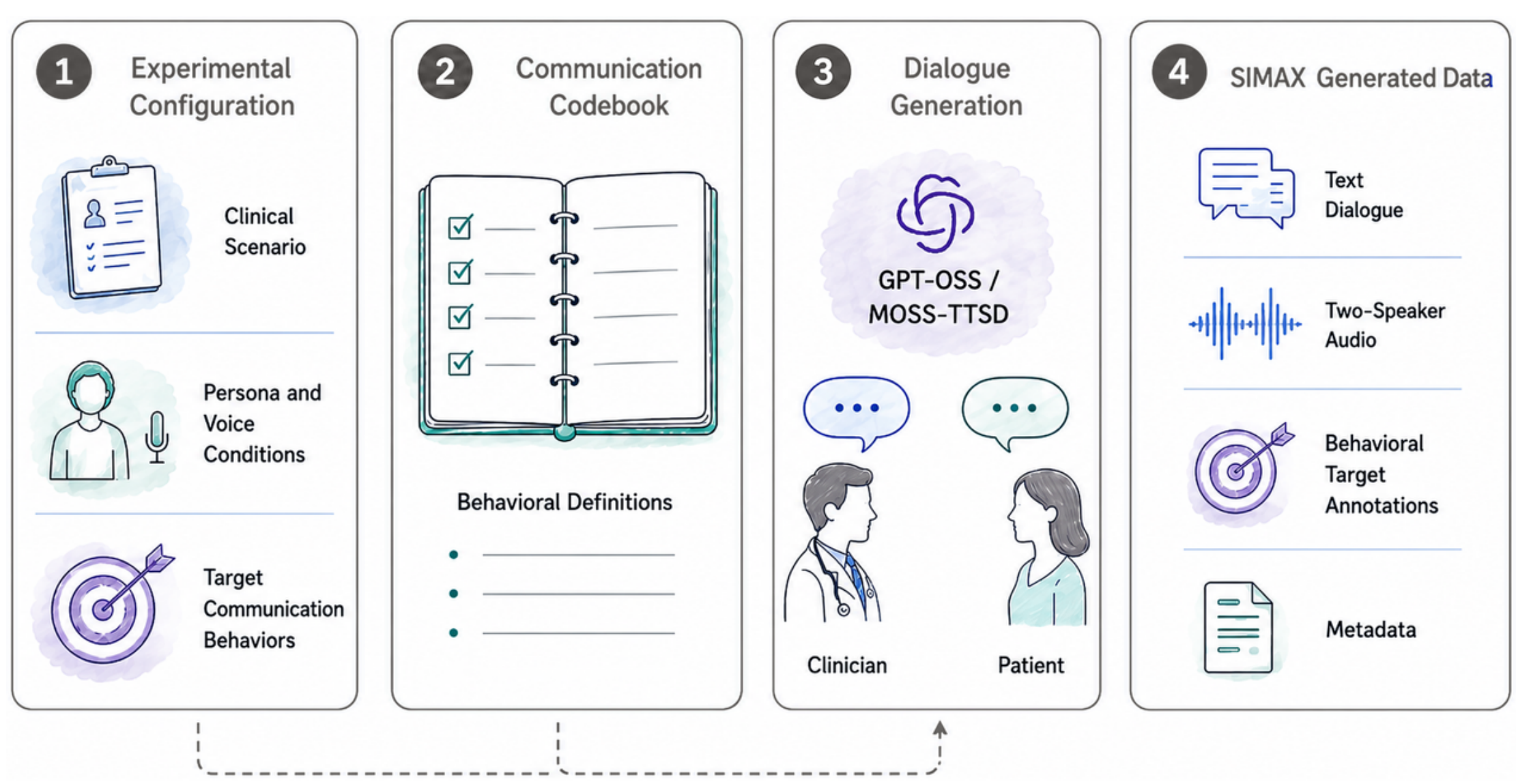

**Figure 1. Overview of the SIMAX framework.** The experimental configuration defines the clinical scenario, persona and voice conditions, and target communication behaviors. Structured communication codebooks translate these behavioral targets into concrete definitions, which are used to guide text dialogue generation by gpt-oss-20B and subsequent audio synthesis by MOSS-TTSD v0.7. SIMAX outputs multimodal records containing text dialogue, two-speaker audio, behavioral target annotations, and related metadata.

*Experimental Configurations and Communication Target Specification*

For each simulated dialogue, the experimental configuration in SIMAX defines three controllable components: the clinical scenario, persona and voice conditions, and target communication behaviors. The clinical scenario specifies the visit context in which the dialogue takes place, including the clinical specialty, visit stage, and contextual background (e.g., birth planning discussion, playground slide injury, or osteoarthritis pain management). Persona and voice conditions specify the individual and vocal characteristics of the clinician and patient, including age, name, gender, voice source, and accent condition. Target communication behaviors specify the intended level of interactional behaviors to be expressed in the dialogue.

To make these target communication behaviors interpretable and reproducible, SIMAX uses two complementary structured communication codebooks to translate them into concrete behavioral standards. We selected the Global Codebook and the WISER Codebook because they capture two important levels of clinical communication coding: the former focuses on overall communication quality and relational interaction, whereas the latter focuses on more specific and countable clinician communication behaviors. Specifically, the Global Codebook captures attentiveness (e.g., consistently responding to the patient's narrative), concern (e.g., addressing patient worries), flow (e.g., naturally guiding the consultation), warmth (e.g., expressing support and friendliness), and respect (e.g., responding to the patient in a nonjudgmental manner), with each dimension defined as an ordinal target on a 1–5 scale[14,15]. The WISER Codebook focuses on empathic responses (e.g., explicitly acknowledging patient emotions), open-ended questions (e.g., inviting the patient to elaborate), and reflective statements (e.g., summarizing the patient's expressed concerns), which are defined as target count ranges[10]. Through this design, each simulated dialogue has a clearly defined clinical context, persona and voice setting, and predefined communication behavior targets.

*Text Dialogue Generation and Audio Synthesis*
During text dialogue generation, SIMAX incorporates clinical scenarios, persona specifications, and codebook-derived behavioral definitions into the generation prompt, as detailed in **Appendix A**. Text dialogues are generated by gpt-oss-20B[16] following a standardized clinical workflow, including greeting, history of present illness, diagnostic reasoning, treatment planning, and summary. The generated text is then synthesized into two-speaker audio using MOSS-TTSD v0.7 to distinguish clinician and patient voices. Clinician reference audio was drawn from either short Common Voice[17] clips (public-short) or longer internally recorded clips (internal-extended), while patient reference audio was selected from Common Voice[17] to match the target accent distribution.

The audio conditions included five English accent variants—African, American, Australian, British, and South Asian—to cover common speech differences across regional and linguistic backgrounds and increase the vocal diversity of simulated clinical dialogues. To better approximate real clinical auditory environments, SIMAX further generated a wet-audio version with background environmental sounds. Specifically, we selected common environmental sounds, such as door sounds, mouse clicks, and keyboard typing, from publicly available sound libraries[18] and mixed them with the original dry audio to simulate environmental sounds that may occur during clinical conversations. Each simulated record was ultimately stored with the dialogue text, dry and wet audio, behavioral target annotations, and metadata including clinical scenario and persona specifications, forming a multimodal input for the downstream evaluation of communication coding systems.

## SIMAX Evaluation

*Evaluation Overview*
We evaluated SIMAX from two complementary dimensions: intrinsic data quality and downstream utility for communication coding systems. Intrinsic data quality assessment examined whether the generated data had high-quality audio and reasonable clinical realism. Downstream utility assessment examined whether the simulated dialogues could be used to test how a communication coding system responds to controlled variation in communication behaviors.

*Intrinsic Data Quality Assessment*
Intrinsic data quality assessment included automated audio quality assessment and human evaluation. Automated metrics included UTMOS[19] and WV-MOS[20] to estimate speech naturalness; both are interpreted on a 1–5 scale, with higher scores indicating better estimated speech naturalness. We used word error rate (WER) and character error rate (CER) to measure transcription errors in the synthesized audio relative to the source text, with lower values indicating higher transcription fidelity. In addition, we used contrastive language-audio pretraining (CLAP) cosine similarity[21] to assess semantic consistency between text and audio; this metric ranges from -1 to 1, with higher values indicating stronger text–audio semantic correspondence.

Human evaluation included two components. First, two expert raters, one native English speaker (E.B.) and one non-native English speaker (L.L.), used a multi-item MOS protocol to assess audio clarity, naturalness, and speaker differentiation, each rated on a 5-point Likert scale. Second, two medical students with patient-facing clinical training experience, D.Y. and R.H., used a 5-point Likert scale to assess the clinical realism of a random sample of 40 records. The detailed human evaluation rubric for intrinsic data quality assessment is provided in **Appendix B**.

*Downstream Utility for Communication Coding Systems*

To assess the downstream utility of SIMAX for evaluating communication coding systems, we used MOSAIC as an example[8]. SIMAX generated dialogues contain predefined behavioral targets derived from the Global Codebook and WISER Codebook, which were used as reference targets for comparison. We input audio-derived transcripts into MOSAIC and obtained communication coding outputs for each dialogue, including Global Codebook ordinal scores and WISER behavior counts. Finally, we compared the MOSAIC outputs against the predefined behavioral targets in SIMAX to examine whether the coding system produced consistent responses to changes in the intensity of communication behaviors. Through this process, SIMAX can serve as controlled evaluation data for benchmarking communication behavior coding in clinical dialogues.

## Results

### Dataset Characteristics and Behavioral Target Distribution

*Dataset Characteristics*

As shown in **Table 1**, SIMAX generated 3,388 simulated clinician–patient dialogues, including 1,801 records in the Global Codebook batch and 1,587 records in the WISER Codebook batch. Each encounter averaged 26.6 turns and 4.99 minutes in duration, with the Global Codebook batch showing slightly higher mean turns and mean duration than the WISER Codebook batch.

| | Global Codebook batch (N=1,801) | WISER Codebook batch (N=1,587) | Overall (N=3,388) |
|---|---|---|---|
| **Clinical Specialty** | | | |
| Obstetrics | 558 (31.0%) | 487 (30.7%) | 1,045 (30.8%) |
| Orthopedics | 1,099 (61.0%) | 975 (61.4%) | 2,074 (61.2%) |
| Rheumatology | 144 (8.0%) | 125 (7.9%) | 269 (7.9%) |
| **Visit Stage** | | | |
| Initial Assessment and Diagnosis | 449 (24.9%) | 411 (25.9%) | 860 (25.4%) |
| Treatment and Care Planning | 373 (20.7%) | 321 (20.2%) | 694 (20.5%) |
| Routine Follow-up and Monitoring | 505 (28.0%) | 442 (27.9%) | 947 (28.0%) |
| Outcome Evaluation and Rehabilitation | 307 (17.0%) | 264 (16.6%) | 571 (16.9%) |
| Acute Flare and Urgent Care | 167 (9.3%) | 149 (9.4%) | 316 (9.3%) |
| **Age** | | | |
| Child (<13) | 290 (16.1%) | 261 (16.4%) | 551 (16.3%) |
| Adolescent (13-17) | 149 (8.3%) | 133 (8.4%) | 282 (8.3%) |
| Adult (18-64) | 819 (45.5%) | 715 (45.1%) | 1,534 (45.3%) |
| Senior (>65) | 543 (30.1%) | 478 (30.1%) | 1,021 (30.1%) |
| **Gender** | | | |
| Female | 965 (53.6%) | 849 (53.5%) | 1,814 (53.5%) |
| Male | 836 (46.4%) | 738 (46.5%) | 1,574 (46.5%) |
| **Voice Source** | | | |
| Public-Short | 883 (49.0%) | 773 (48.7%) | 1,656 (48.9%) |
| Internal-Extended | 918 (51.0%) | 814 (51.3%) | 1,732 (51.1%) |
| **Accent Condition** | | | |
| African | 334 (18.5%) | 287 (18.1%) | 621 (18.3%) |
| American | 357 (19.8%) | 316 (19.9%) | 673 (19.9%) |
| Australian | 389 (21.6%) | 340 (21.4%) | 729 (21.5%) |
| British | 365 (20.3%) | 322 (20.3%) | 687 (20.3%) |
| South Asian | 356 (19.8%) | 322 (20.3%) | 678 (20.0%) |
| **Encounter Characteristics** | | | |
| Mean Turns per Encounter | 29.1 | 23.8 | 26.6 |
| Mean Duration (minutes) | 5.43 | 4.49 | 4.99 |

**Table 1. Overview of SIMAX generated clinician–patient dialogues.** Distributions are shown by codebook batch across clinical specialty, visit stage, persona characteristics, voice conditions, and encounter characteristics. Public-short refers to short publicly available reference voice clips selected from Common Voice, whereas internal-extended refers to longer internally recorded clinician reference voice clips.

The dataset covered three clinical specialties: obstetrics, orthopedics, and rheumatology. Orthopedics accounted for the largest proportion of records (2,074/3,388, 61.2%), followed by obstetrics (1,045/3,388, 30.8%) and rheumatology (269/3,388, 7.9%). In terms of visit stage, the simulated dialogues covered five clinical contexts: initial assessment and diagnosis, treatment and care planning, routine follow-up and monitoring, outcome evaluation and rehabilitation, and acute flare and urgent care. Routine follow-up and monitoring (947/3,388, 28.0%) and initial assessment and diagnosis (860/3,388, 25.4%) were the most common visit stages. The two codebook batches showed similar distributions across clinical specialties and visit stages.

The dataset also included controlled persona and voice conditions. Adults represented the largest age group (1,534/3,388, 45.3%), followed by seniors (1,021/3,388, 30.1%), children (551/3,388, 16.3%), and adolescents (282/3,388, 8.3%). Gender distribution was relatively balanced, with female and male personas accounting for 53.5% and 46.5% of records, respectively. The two reference audio sources, public-short and internal-extended, were similarly represented, accounting for 48.9% and 51.1% of records. Accent conditions covered African, American, Australian, British, and South Asian variants, with each category accounting for 18.3% to 21.5% of records.

*Behavioral Target Distribution*

For behavioral target specification, SIMAX adopted a balanced target allocation strategy to ensure systematic coverage of different communication behavior intensities in the simulated data. For the Global Codebook batch, the predefined target levels for each communication dimension were 1, 3, and 5, representing low, moderate, and high levels of overall communication performance, respectively. For the WISER Codebook batch, the predefined target ranges for each countable communication behavior were [1,2], (2,4], and (4,7], representing low, moderate, and high frequencies of behavior occurrence, respectively. Through this design, SIMAX covers different types and intensities of communication behavior targets under controlled conditions, providing a basis for subsequently evaluating whether the communication coding system can respond to predefined behavioral variation.

## Intrinsic Data Quality of SIMAX Generated Records

### *Automated Audio Quality Assessment*

Automated audio quality assessment results are shown in **Figure 2**. Mean UTMOS and WV-MOS scores were 3.03 and 2.61, respectively, on a 1–5 scale, indicating reasonable speech naturalness in the generated audio. Transcription results showed low error rates, with overall WER and CER of 0.07 and 0.05, respectively, indicating high content fidelity relative to the source text. For text–audio semantic consistency, the overall CLAP cosine similarity was 0.41, indicating a positive semantic correspondence between the generated audio and the source dialogue text. The Global Codebook batch and WISER Codebook batch showed only small differences across automated metrics, indicating generally stable audio quality across the two generated batches.

a. Speech Naturalness

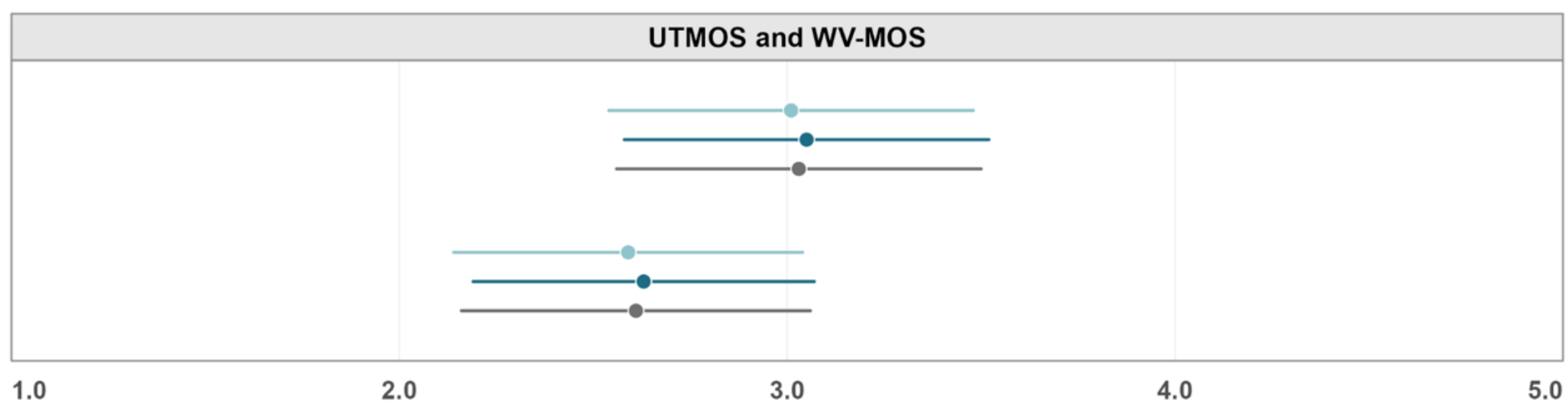

b. Transcription Error

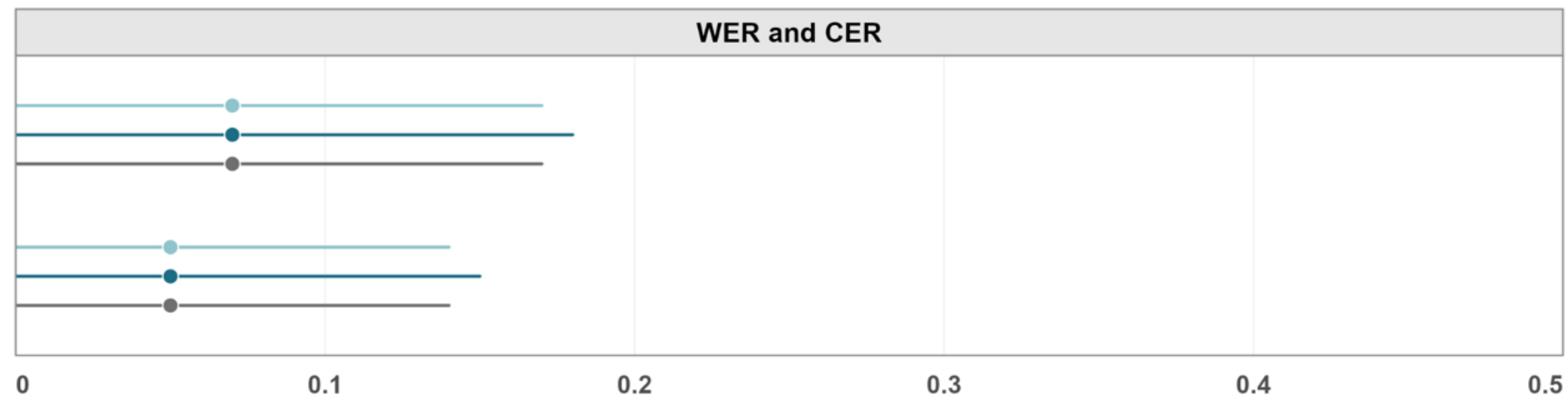

c. Semantic Consistency

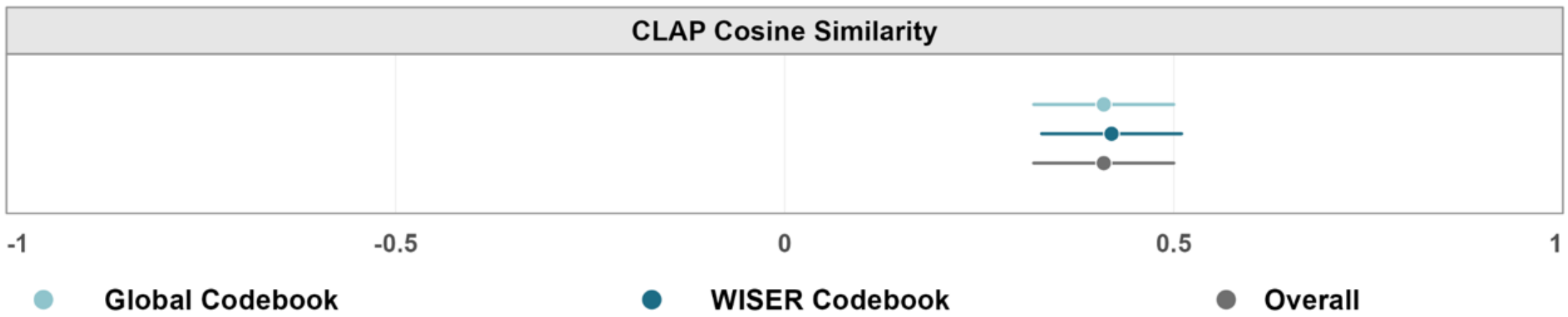

**Figure 2. Automated audio quality assessment of SIMAX generated dialogues.** (a) Speech naturalness assessment: the upper three lines represent UTMOS results, and the lower three lines represent WV-MOS results. (b) Transcription error assessment: the upper three lines represent WER results, and the lower three lines represent CER results. (c) Text–audio semantic consistency assessment using CLAP cosine similarity. The three lines within each group correspond to the Global Codebook batch, WISER Codebook batch, and overall dataset, respectively.

*Human Evaluation*

Human evaluation results are shown in **Figure 3**. Overall, SIMAX generated audio achieved a median MOS of 4.67, suggesting good clarity, naturalness, and speaker differentiation on a 5-point scale. When stratified by accent condition, the American group had the highest median MOS at 4.83; the Australian, South Asian, and British groups each had a median MOS of 4.67, while the African group was relatively lower at 4.50, with modest overall differences across groups. By clinical specialty, obstetrics had the highest median MOS at 4.83, followed by rheumatology at 4.67 and orthopedics at 4.50. For clinical realism, the overall median score was 3.00, suggesting a moderate level of clinical realism in the evaluated scenarios on a 5-point scale. When stratified by accent condition, the South Asian group had the highest median clinical realism score at 4.00. By clinical specialty, obstetrics and rheumatology had median scores of 3.75 and 3.50, respectively, both higher than orthopedics at 2.50, suggesting some variation in clinical realism across specialties.

**Figure 3. Human evaluation of SIMAX generated dialogues.** (a) MOS score: overall MOS scores of SIMAX generated audio are shown and stratified by clinical specialty and accent condition. MOS was rated on a 5-point Likert scale based on clarity, naturalness, and speaker differentiation. (b) Clinical realism score: overall clinical realism scores of SIMAX generated dialogues are shown and stratified by clinical specialty and accent condition. Clinical realism was rated on a 5-point Likert scale based on clinical plausibility and resemblance to real outpatient interactions.

### Downstream Utility for Communication Coding Systems

To assess the downstream utility of SIMAX for evaluating communication coding systems, we input the audio-derived transcripts of SIMAX generated dialogues into MOSAIC and analyzed whether its outputs changed in response to communication behavior targets predefined in SIMAX. As shown in **Figure 4**, MOSAIC responses were not consistent across different types of behavioral targets. In the Global Codebook batch, MOSAIC scores did not show a clear monotonic increasing trend across the three SIMAX predefined target level groups (1, 3, and 5). In the WISER Codebook batch, the behavior counts detected by MOSAIC generally increased as the SIMAX predefined count ranges increased ([1,2], (2,4], and (4,7]). Overall, these results suggest that SIMAX generated data can serve as a controlled evaluation resource for preliminary examination of the response patterns of communication coding systems across different codebook types and communication behavior dimensions, while also helping to identify insufficient sensitivity in certain dimensions.

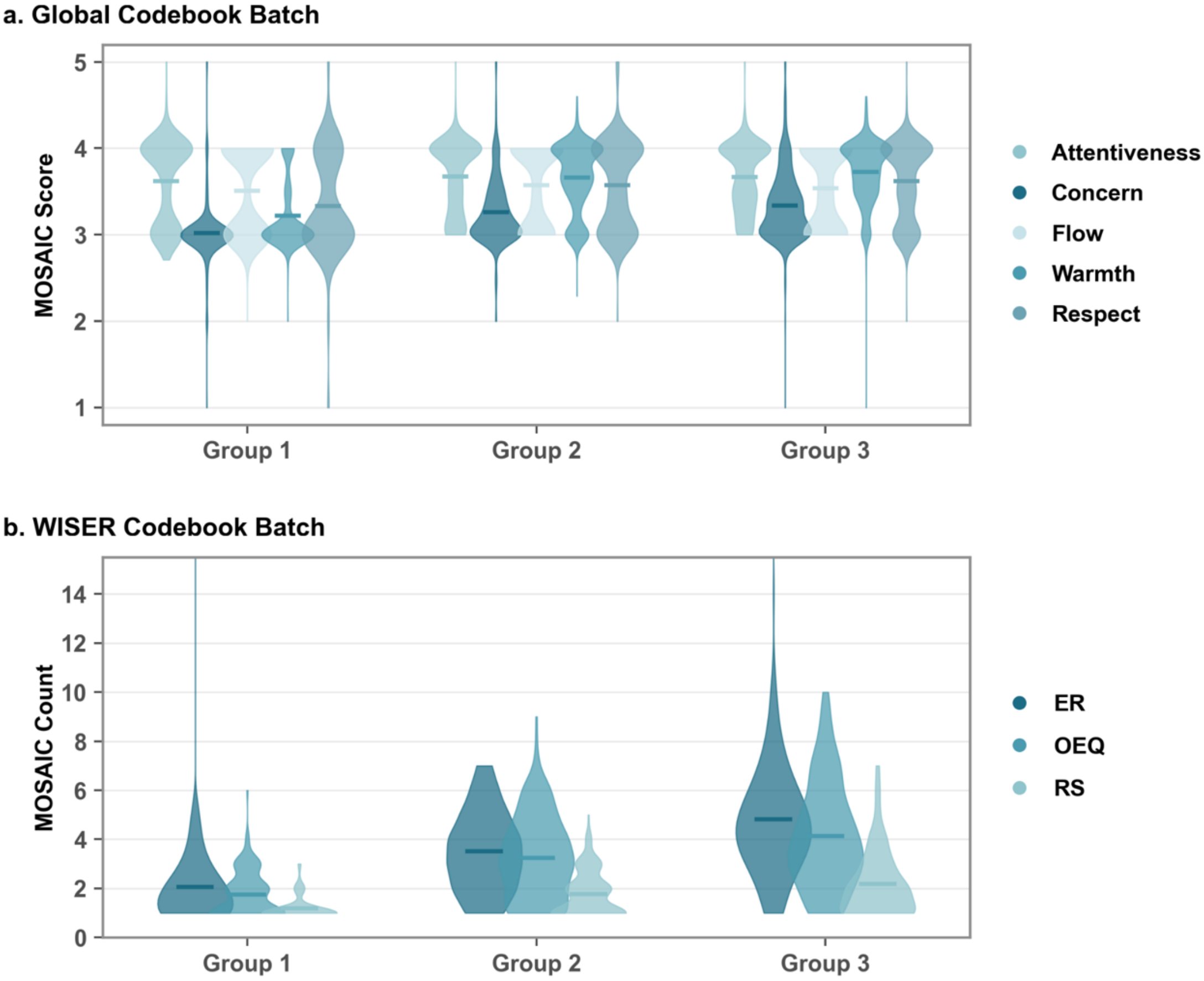

**Figure 4. Downstream utility assessment comparing SIMAX predefined behavioral targets and MOSAIC communication coding system outputs.** (a) Global Codebook batch: ordinal MOSAIC scores for the five Global Codebook dimensions are shown after min–max rescaling and grouped by SIMAX predefined target level groups. The predefined target level groups were 1, 3, and 5, representing low, moderate, and high levels of overall communication behavior targets, respectively. (b) WISER Codebook batch: MOSAIC counts for the three WISER Codebook behaviors are shown and grouped by SIMAX predefined target count range groups. The predefined target count range groups were [1,2], (2,4], and (4,7], representing low, moderate, and high frequencies of communication behavior targets, respectively.

## Discussion

In this study, we developed SIMAX to provide controlled and reproducible clinical dialogue data for communication coding systems. By generating large-scale simulated clinician–patient dialogues with predefined communication behavior targets and reference behavioral annotations, SIMAX enables coding systems to be systematically evaluated under known behavioral conditions, rather than relying solely on limited real-world samples. This data foundation supports clearer assessment of how coding systems respond to variations in communication behavior type and intensity, thereby facilitating the development, validation, and refinement of communication coding systems. Beyond this study, SIMAX is intended as a reusable resource: other users can apply it to their own clinical scenarios and behavioral targets to benchmark communication coding systems without requiring protected real-world dialogues.

We evaluated SIMAX across two dimensions: intrinsic data quality and downstream utility for communication coding systems. For intrinsic data quality, SIMAX generated records showed acceptable usability in terms of audio quality and clinical realism. Automated audio assessment metrics supported speech naturalness, transcription fidelity, and semantic consistency between text and audio, while human evaluation further assessed clarity, naturalness, and speaker differentiation. Clinical realism ratings suggested that the simulated dialogues reflected plausible outpatient interactions in the evaluated scenarios.

To assess the downstream utility of SIMAX, we used the audio-derived transcripts of SIMAX generated dialogues as test inputs to evaluate MOSAIC and compared the MOSAIC outputs against the predefined behavioral targets. The results showed that MOSAIC responses were not fully consistent across different types of behavioral targets. In the Global Codebook batch, MOSAIC scores did not show a clear monotonic increasing trend as the predefined target scores increased; by contrast, in the WISER Codebook batch, the behavior counts detected by MOSAIC generally increased as the predefined count ranges increased. These findings suggest that SIMAX generated data can be used to preliminarily examine the response patterns of communication coding systems across different codebook types and communication behavior dimensions, while also identifying insufficient sensitivity in certain dimensions.

This study has several limitations. First, SIMAX generates synthetic clinical dialogues and should therefore be viewed as a controlled evaluation resource rather than a direct substitute for external validation using real clinical dialogues[22]. Although the simulated dialogues are constructed based on clinical scenarios, they may not fully capture the complexity of real clinical interactions[23]. In addition, the average duration of SIMAX generated dialogues is approximately 5 minutes, which is shorter than the complete clinical exchanges in real outpatient settings, which are typically around 15-20 minutes. Second, behavioral control in the current framework is primarily implemented through communication codebooks, which makes the manipulation of communication behaviors reproducible, but the effectiveness of control varies across different types of behavioral targets. In particular, in the Global Codebook batch, MOSAIC outputs did not show a clear monotonic increasing trend across the predefined target levels. Third, although SIMAX supports environmental audio, the current version does not fully model speech overlap, clinician or patient interruptions, and other phenomena commonly observed in real-world dialogues. Fourth, the downstream evaluation in this study used audio-derived transcripts; therefore, the quality of generated audio may affect automatic transcription and the stability of subsequent communication coding, but this relationship was not systematically evaluated in the current study. Finally, this study did not directly compare SIMAX generated dialogues with human-coded real-world clinical dialogues.

Future work should improve the clinical realism and generalizability of SIMAX. One important direction is to align simulated dialogues with deidentified real clinical recordings, where permitted by ethical and legal frameworks, to quantify differences between them under the same communication coding system and use these differences to calibrate dialogue flow and behavior control strategies. SIMAX should also be extended to more clinical specialties and care settings, especially primary care and telehealth settings where ADS are increasingly being adopted[5]. In addition, multilingual expansion will be important to support the evaluation of communication coding systems across broader patient populations and clinical environments[24,25,26].

## Supplementary Appendix

### Appendix A: Prompt Templates for SIMAX Dialogue Generation

SIMAX uses a two-stage pipeline for text dialogue generation. In the first stage, the LLM generates structured information for the clinical scenario, clinician, and patient. In the second stage, this information is combined with codebook-derived behavioral targets to generate the clinician–patient dialogue. The full prompt templates used in both stages are provided below.

*Clinical Scenario Generation Prompt Template*

> Generate a detailed clinical scenario:
>
> {clinical_specialty}
> {visit_stage}
> {contextual_background}
>
> **Output Format:**
> Please fill in the following JSON template. Output ONLY valid JSON without any markdown code blocks or additional text.
>
> ```
> {
>   "demographics": "",
>   "chief_complaint": "",
>   "present_illness_history": "",
>   "past_medical_history": "",
>   "family_history": ""
> }
> ```

*Clinician Profile Generation Prompt Template*

> Generate a clinician profile for a clinician–patient dialogue simulation according to the following criteria:
>
> {clinical_scenario}
> {global_scores}

{wiser_counts}

**Output Format:**
Please fill in the following JSON template. Output ONLY valid JSON without any markdown code blocks or additional text.

```
{
  "name": "",
  "gender": "",
  "age": "",
  "title": "",
  "personality": "",
  "behavior": ""
}
```

*Patient Profile Generation Prompt Template*

Generate a patient profile for a clinician–patient dialogue simulation according to the following criteria:

{clinical_scenario}
{global_scores}
{wiser_counts}

**Output Format:**
Please fill in the following JSON template. Output ONLY valid JSON without any markdown code blocks or additional text.

```
{
  "name": "",
  "gender": "",
  "age": "",
  "personality": "",
  "behavior": ""
}
```

*Dialogue Generation Prompt Template*

> Generate a dialogue between a clinician and a patient according to the following information:
>
> Clinical Scenario: {clinical_scenario}
> Clinician: {clinician_profile}
> Patient: {patient_profile}
>
> The following codebook-grounded behavioral targets MUST be reflected in the dialogue content.
>
> {behavioral_targets}
>
> The dialogue should be realistic and follow the clinical workflow.
> The dialogue should have between {min_turns} and {max_turns} turns.
>
> The dialogue must follow this five-stage sequence in order:
> 1. Greeting
> 2. History of Present Illness
> 3. Diagnostic Reasoning
> 4. Treatment Planning
> 5. Summary
>
> Crucially, to ensure natural-sounding speech synthesis:
> 1. Include natural filler words and hesitations (e.g., 'um', 'uh', 'well', 'hmm', 'ah') where appropriate, especially for the patient who might be nervous.
> 2. Do NOT use stage directions or sound descriptions (e.g., pauses, coughs, [sighs], or (silence)).
> 3. Keep the filler words moderate and natural; do not overuse them.
>
> Format the output strictly following the MOSS-TTSD input format:
> 1. Each turn must be on a new line.
> 2. Start each line with '[S1]' for the Clinician and '[S2]' for the Patient.
> 3. The Clinician ([S1]) must speak first.
> 4. Do not include any other text, markdown, or JSON formatting.

## Appendix B: Human Evaluation Rubric for Intrinsic Data Quality Assessment

*MOS Rating*

**Rating 1 (Bad)**
**Clarity:** Speech is largely unintelligible; noise or distortion severely interferes with understanding.
**Naturalness:** Completely mechanical; pacing and pauses feel unnatural and robotic.
**Speaker Differentiation:** Voices are identical; impossible to distinguish roles by audio.

**Rating 2 (Poor)**
**Clarity:** Speech is understandable only with effort; audio issues frequently interfere with comprehension.
**Naturalness:** Noticeably synthetic; rhythm and phrasing are often awkward or poorly timed.
**Speaker Differentiation:** Voices are very similar; requires intense concentration to tell them apart.

**Rating 3 (Fair)**
**Clarity:** Speech is mostly understandable; some audio issues are noticeable but only occasionally disrupt comprehension.
**Naturalness:** Understandable and fairly smooth, but still clearly synthetic in rhythm or phrasing.
**Speaker Differentiation:** Voices are distinct enough to follow, though they may share similar tonal qualities.

**Rating 4 (Good)**
**Clarity:** Speech is clear and easy to follow; minor audio issues are present but do not meaningfully affect understanding.
**Naturalness:** Mostly natural rhythm and pacing, with only minor awkward pauses or phrasing.
**Speaker Differentiation:** Good separation; differences in pitch or speaking style are obvious.

**Rating 5 (Excellent)**

**Clarity:** Speech is exceptionally clear throughout; the audio is fully intelligible with no meaningful barriers to comprehension.
**Naturalness:** Very natural rhythm, pacing, and phrasing; sounds human-like.
**Speaker Differentiation:** Excellent separation; voices have completely different characteristics (e.g., age, gender, accent) with zero ambiguity.

*Clinical Realism Rating*

**Rating 1 (Bad)**
Contains critical medical errors or nonsensical content; clinically dangerous or absurd.

**Rating 2 (Poor)**
Basic logic is present, but contains frequent misuse of terminology or unnatural questioning.

**Rating 3 (Fair)**
Medically correct but rigid; sounds like a student reading a textbook rather than a practitioner.

**Rating 4 (Good)**
Realistic flow with correct terminology; represents a standard, competent clinical encounter.

**Rating 5 (Excellent)**
Highly realistic; captures the nuance, shorthand, and complex reasoning of an experienced clinician.